\pdfoutput=1
\documentclass[11pt]{article}

\usepackage{ACL2023}

\usepackage{times}
\usepackage{latexsym}
\usepackage{booktabs}
\usepackage{amsmath}
\usepackage{ulem}
\usepackage[T1]{fontenc}
\usepackage{algorithm}
\usepackage{algorithmic}
\usepackage{enumitem}
\usepackage[utf8]{inputenc}
\newcommand{\ie}{\textit{i.e.}}
\newcommand{\eg}{\textit{e.g.}}

\newcommand{\TE}{T\!E}

\newcommand{\IE}{I\!E}
\newcommand{\NDE}{N\!D\!E}

\newcommand{\TDE}{T\!D\!E}
\newcommand{\TIE}{T\!I\!E}

\usepackage{multirow}
\usepackage{multicol}
\usepackage{amsmath, amssymb}
\usepackage{graphicx} % \resizebox
\usepackage{booktabs} % beautify formating
\usepackage{makecell}
\usepackage{colortbl}
\usepackage{tabularx}
\definecolor{mygray}{gray}{.95}
\usepackage{microtype}
\usepackage{xcolor} 
\usepackage{pifont}

\newcommand{\yes}{\color{green!60!black}\ding{51}}
\newcommand{\no}{\color{red!60!black}\ding{55}}
\usepackage{inconsolata}

\title{DINER: Debiasing Aspect-based Sentiment Analysis with Multi-variable Causal Inference}

\author{Jialong Wu\thanks{~~Equal Contribution.}$^{\spadesuit}$$^{\diamondsuit*}$\hspace{1.5mm}
Linhai Zhang$^{\spadesuit}$$^{\diamondsuit*}$\hspace{1.5mm}
Deyu Zhou$^{\spadesuit}$$^\diamondsuit$\thanks{~~Corresponding Author.}\hspace{1.5mm}
Guoqiang Xu$^{\heartsuit}$\\
        $^{\spadesuit}$\hspace{0.5mm}School of Computer Science and Engineering, Southeast University, Nanjing, China \\
        $^\diamondsuit$Key Laboratory of New Generation Artificial Intelligence Technology and Its \\
Interdisciplinary Applications (Southeast University), Ministry of Education, China
\\ $^{\heartsuit}$\hspace{0.5mm}SANY Group Co., Ltd. \\
        \texttt{\{jialongwu, lzhang472, d.zhou\}@seu.edu.cn} \\
        \texttt{xuguoqiang-2012@hotmail.com}
        } 

\begin{document}
\maketitle

\begin{abstract}
Though notable progress has been made, neural-based aspect-based sentiment analysis (ABSA) models are prone to learn spurious correlations from annotation biases, resulting in poor robustness on adversarial data transformations.
Among the debiasing solutions, causal inference-based methods have attracted much research attention, which can be mainly categorized into causal intervention methods and counterfactual reasoning methods. 
However, most of the present debiasing methods focus on single-variable causal inference, which is not suitable for ABSA with two input variables (\textit{the target aspect} and \textit{the review}).  
In this paper, we propose a novel framework based on multi-variable causal inference for debiasing ABSA. In this framework, different types of biases are tackled based on different causal intervention methods. For the review branch, the bias is modeled as indirect confounding from context, where backdoor adjustment intervention is employed for debiasing.
For the aspect branch, the bias is described as a direct correlation with labels, where counterfactual reasoning is adopted for debiasing.
Extensive experiments demonstrate the effectiveness of the proposed method compared to various baselines on the two widely used real-world aspect robustness test set datasets. \footnote{ Our code
and results will be available at \url{https://github.com/callanwu/DINER}.
}
\end{abstract}

\section{Introduction}
Aspect-Based Sentiment Analysis~(ABSA) aims to classify the polarity of the sentiment ($\eg$, positive, negative, or neutral) towards a specific aspect of a sentence (\eg, \textit{bugers} in the review ``\textit{Tasty bugers, and crispy fries.}'')~\cite{hu2004mining,jiang-etal-2011-target,vo2015target,zhang2016gated,zhang2022survey}.
Most ABSA methods solve the task as an input-output mapping problem based on high-capacity neural networks and pre-trained language models~\cite{wang-etal-2018-target,huang-carley-2018-parameterized,bai2020investigating}.  
Though remarkable progress has been made, it is demonstrated that these state-of-the-art models are not robust in data transformation where simply reversing the polarity of the target results in over 20\% drop in accuracy~\cite{xing-etal-2020-tasty}.

\begin{figure}[t]
    \centering
    \includegraphics[width=0.49\textwidth]{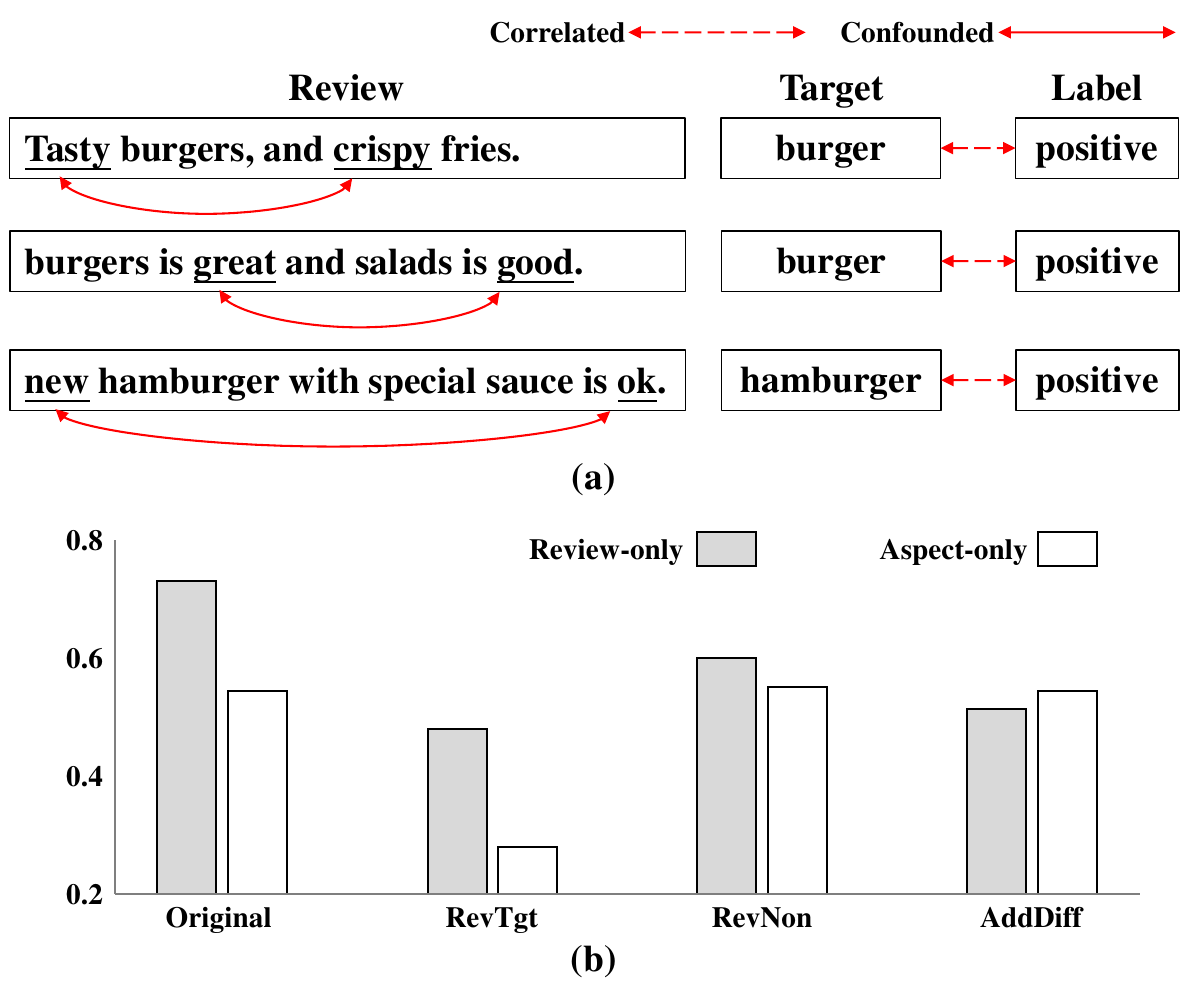}
    \caption{(a) Examples are taken from the SemEval 2014 Restaurant test set. (b) \textsc{RevTgt} denotes reversing the polarity of the target aspect, \textsc{RevNon} denotes reversing the polarity of the non-target aspect, and \textsc{AddDiff} denotes adding another non-target aspect with different polarity.}
    \label{fig:example}
\end{figure}

A reasonable explanation is that neural networks trained with the Stochastic Gradient Descent algorithm are vulnerable to annotation biases and learn the shortcuts instead of the underlying task~\cite{xing-etal-2020-tasty}. 
As shown in Figure~\ref{fig:example} (a), over 50.0\% of targets have only one kind of polarity label in the widely used SemEval 2014 Laptop and Restaurant datasets~\cite{pontiki-etal-2014-semeval}. 
For 83.9\% and 79.6\% instances in the test sets, the sentiments of the target aspect and all non-target aspects are the same. 
Therefore, it is easy for end-to-end neural models to learn such spurious correlations and make predictions solely based on target aspects or sentiment words describing non-target aspects. 

To avoid learning spurious correlations, recent methods focus on debiasing, which can be categorized into argumentation-based methods~\cite{wei-zou-2019-eda, 10.1145/3459637.3482078}, reweight training-based methods~\cite{schuster-etal-2019-towards, karimi-mahabadi-etal-2020-end} and causal inference-based methods~\cite{niu2021counterfactual,liu2022contextual}. 
Among them, causal inference attracts much research interest for its theoretical-granted property and little modification to the existing learning paradigm. 
\citet{niu2021counterfactual} proposed a debiasing method for the language bias in the vision question answering task by performing counterfactual reasoning.
\citet{liu2022contextual} employed backdoor adjustment-based intervention for mitigating the context bias in object detection.
Recent attempts have been made to solve various biases in natural language processing tasks, including natural language understanding~\cite{tian2022debiasing}, implicit sentiment analysis~\cite{wang-etal-2022-causal}, and fact verification~\cite{xu-etal-2023-counterfactual}.

However, most causal inference-based debiasing methods are based on single-variable causal inference, which is not appropriate for ABSA with two input variables.
As shown in Figure~\ref{fig:example} (a), there are two types of biases in ABSA. 
The target aspects $A$ are often directly correlated with the polarity labels $L$, while the sentiment words for targets in the review $R$ are often indirectly confounded with the non-targets $C$. 
To further investigate the difference between aspect-related biases and review-related biases, a simple experiment is conducted by training two probing models with only review or aspect as input.
As shown in Figure~\ref{fig:example} (b), the aspect-only model has similar performances on the original and the adversarial test set except \textsc{RevTgt} where spurious correlations learned in the training set are flipped, while the review-only model performs differently on four test variants. 
It might suggest that the biases in the aspect branch are direct and simple, while the biases in the review branch are indirect and complicated, which poses a challenge. 

To tackle the above challenge, we propose  \textbf{D}ebias \textbf{IN} Asp\textbf{E}ct and \textbf{R}eview (\textbf{DINER}) based multi-variable causal inference for debiasing ABSA.
To be more specific, as illustrated in Figure~\ref{fig:SCM}, the unbiased prediction is obtained by calculating the total indirect effect of the target aspect and the review of the polarity label, which is further decomposed and estimated by the hybrid causal intervention method.
For the $R \rightarrow L$ branch, a backdoor adjustment intervention is employed to mitigate the indirect confounding between the target sentiment words in the review and the context. 
For the $A \rightarrow L$ branch, a counterfactual reasoning intervention is employed to remove the direct correlation between the target and the label.
Extensive experiments on two widely used real-world robustness test benchmark datasets show the effectiveness of our framework. 

Overall, our contributions can be summarized as follows:
\begin{itemize}
    \item A novel framework is proposed for debiasing ASBA based on multi-variable causal inference. As far as we know, we are the first to uncover and analyze the bias problem in ABSA using multi-variable causal inference.
    \item 
    A hybrid intervention method is constructed by combining backdoor adjustment and counterfactual reasoning. 
    \item The detailed evaluation demonstrates that the proposed method empirically advances the state-of-the-art baselines. 
\end{itemize}

\section{Related Work}
Our work is mainly related to two lines of research, described as follows. 

\subsection{Aspect-Based Sentiment Analysis}

ABSA has garnered significant research attention in recent years. 
Early works focus on feature engineering with manual-construction sentiment lexicons and syntactic features, and rule-based classifiers are adopted to make predictions~\cite {jiang-etal-2011-target,kiritchenko-etal-2014-nrc}.
With the development of neural networks and word embedding techniques, neural-based models have dominated the area with architectures such as LSTM, CNN, Attention mechanisms, Capsule Network~\cite{tang-etal-2016-effective,wang-etal-2016-attention,xue-li-2018-aspect,jiang-etal-2019-challenge}.
Recent advances in pre-trained language models such as BERT~\cite{devlin-etal-2019-bert} have shifted the paradigm again~\cite{zhang2022survey}, where most recent models take pre-trained models as backbones~\cite{xu-etal-2019-bert,hou2021graph,cao2022aspect}.
However, ABSA still faces challenges on robustness datasets, and it is precisely such tasks that our approach targets.

\subsection{Causal Inference-based Debiasing}
Causal inference~\cite{pearl1995causal,pearl2009causal} has been widely employed for debiasing in various fields, including computer vision, recommendation, and natural language processing~\cite{niu2021counterfactual,zhang2021causal,tian2022debiasing}.
The main methods employed consist of counterfactual reasoning and causal intervention.
\citet{niu2021counterfactual} proposed to remove the language bias in vision question answering by subtracting the results of a counterfactual language-only model from the results of a vanilla language-vision model.
Following this work, counterfactual reasoning is widely applied to debiasing the spurious correlation between input and label in tasks including natural language understanding~\cite{tian2022debiasing}, machine reading comprehension~\cite{guo-etal-2023-counterfactual,zhu-etal-2023-causal} and fact verification~\cite{xu-etal-2023-counterfactual}.
\citet{liu2022contextual} proposed to de-confound the object from the context in object detection with backdoor adjustment, where an inverse probability weight approximation is made to estimate the \textit{do}-operator.
Another way to estimate the \textit{do}-operator is known as normalized weighted geometrical mean (NWGM), which is firstly adopted in image caption by \citet{Liu_2022_CVPR}.
Following this line of work, backdoor adjustment-based debiasing has widely been explored in tasks including named entity recognition~\cite{zhang-etal-2021-de} and multi-modal fake news detection~\cite{chen2023causal}. 
Some methods also employ other causal inference techniques, including instrument variable~\cite{wang-etal-2022-causal} and colliding effects~\cite{zheng-etal-2022-distilling}.
However, most of the present debiasing methods focus on debiasing a single input variable, while we are the first to debias two input variables in ABSA simultaneously.
%our method focuses on simultaneously debiasing two input variables in ABSA.

\section{Methods}
In this section, we will introduce the proposed method, \textbf{DINER}, in detail. 
First, we will define the Structural Causal Model (SCM) of ABSA and derive the formula of causal effect step by step.
Then, we will formulate how to estimate the components in the causal effect formula with backdoor adjustment and counterfactual reasoning.
Finally, we will introduce the training and inference processes.

\subsection{Structural Causal Model of ABSA}

\begin{figure}[t]
    \centering
    \includegraphics[width=0.49\textwidth]{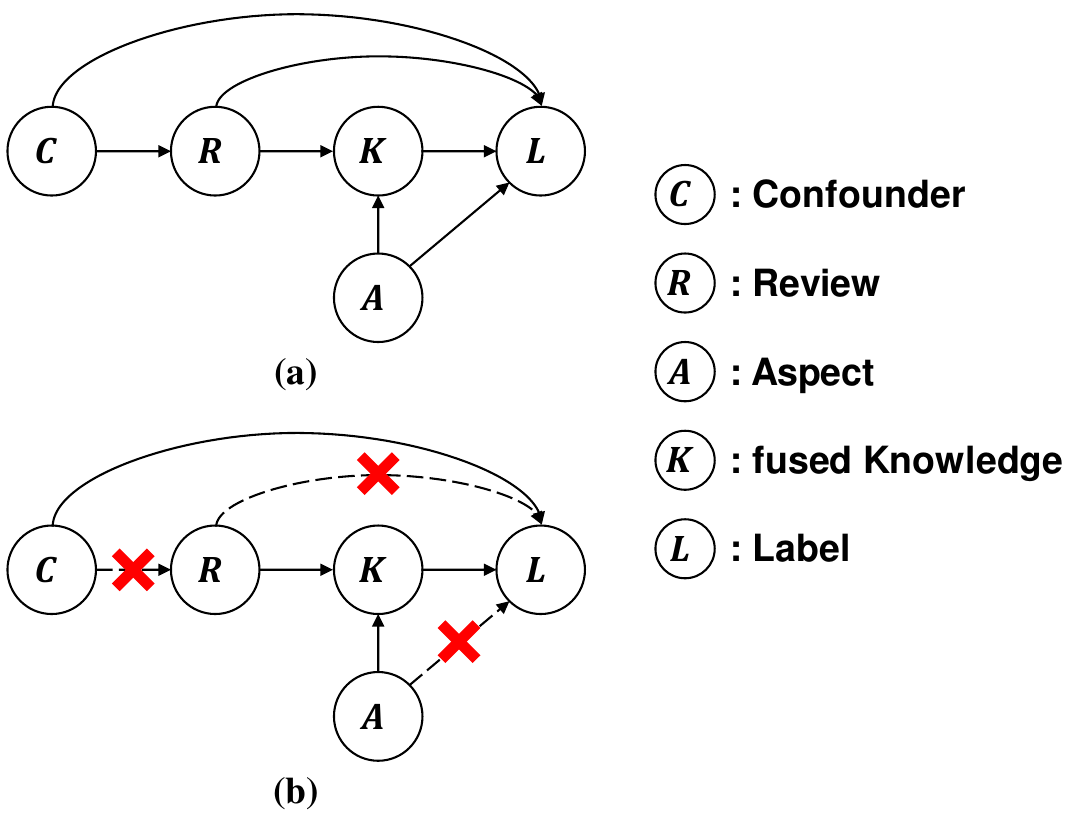}
    \caption{(a) SCM of the proposed method. (b) The desired situation for ABSA, the dotted line means the causalities are blocked.}
    \label{fig:SCM}
\end{figure}

The SCM of ABSA, which is formulated as a directed acyclic graph, is shown in Figure~\ref{fig:SCM} (a). 
The nodes in the SCM denote causal variables, and the edges denote causalities between two nodes ($\eg$, $X \rightarrow Y$ means $X$ causes $Y$).
Then we will discuss the rationale behind how this SCM is built:
\begin{itemize}[leftmargin=*]
    \item $R \rightarrow K \leftarrow A$. The prediction of ABSA is dependent on both review $R$ and aspect $A$. Therefore, a fused knowledge node $K$ is caused by both $R$ and $A$.
    \item $K \rightarrow L$. The label $L$ is caused by the fused knowledge $K$, which is the desired causal effect of ABSA.
    \item $R \rightarrow L \leftarrow A$. The label $L$ is also directly affected by review $R$ and aspect $A$, where the spurious correlation comes from and should be removed.
    \item $C \rightarrow R$ and $C \rightarrow L$. The confounder $C$ (the prior context knowledge) caused $R$ and $L$ simultaneously, where the annotation biases come from. For example, most reviews contain positive descriptions for multiple types of food, which will encourage the model to make predictions without identifying the target.
\end{itemize}
It is worth noticing that we do not add the edge $C \rightarrow A$ or $R \rightarrow A$.
Because we believe the choice of aspect $A$ is made by the annotators and not restricted by the context $C$ or review $R$.

With the SCM defined, we can derive the formula of causal effect.
As shown in Figure~\ref{fig:SCM} (b), the desired situation for ABSA is that the edges that bring biases are all blocked, and the prediction is based on aspect $A$ and review $R$ solely through the fused knowledge $K$.
With the language of causal inference, the prediction should be made on:
\begin{equation}
    \TIE_{a,r} = \TE_{a,r} - \NDE_r - \NDE_a + \IE_{a,r}
\end{equation}
where $\TIE_{a,r}$ denotes the Total Indirect Effect ($\TIE$) from $A$ and $R$ on $L$, $\TE_{a,r}$ denotes the Total Effect ($\TE$), $\NDE$ denotes the Natural Direct Effect ($\NDE$), and $\IE_{a,r}$ denotes the Interaction Effect ($\IE$) between $A$ and $R$.
The total effect $\TE$ contains all causal effects from $A$ and $R$ on $L$, inducing the biases, while the natural direct effect ($\NDE$) only measures the direct causal effect between two variables, which can be regarded as the bias-only effect.
Therefore, subtracting $\NDE_a$ and $\NDE_r$ from $\TE_{a,r}$ will results in the unbiased causal effect from $A$ and $R$ on $L$, which is the total indirect effect $\TIE_{a,r}$.
It is worth noticing that since there is no causality between $A$ and $R$, the value of the interaction effect $\IE_{a,r}$ can be set to 0.

Based on the defintion of $\TE$ and $\NDE$~\cite{niu2021counterfactual}:
\begin{align}
    \TE_{a,r} &= L_{a,r,k} - L_{a^*,r^*,k^*} \\
    \NDE_a &= L_{a,r^*,k^*} - L_{a^*,r^*,k^*} \\
    \NDE_r &=  L_{a^*,r,k^*} - L_{a^*,r^*,k^*}
\end{align}
where $L$ denotes the prediction and $x^{*}$ denotes variable $x$ is set to be void, we can have:
\begin{equation}
\begin{split} 
    \TIE_{a,r} &= L_{a,r,k} -  L_{a^*,r,k^*} -
    L_{a,r^*,k^*} +
    L_{a^*,r^*,k^*} \\
    &= \TIE_{a,r^{'}} - \NDE_{a}
\end{split} 
\label{eq:tie}
\end{equation}
where $r^{'}$ denotes the debiased review, obtained after the process of deconfounding.

\subsection{Deconfounding the Review Branch with Backdoor Adjustment}
\label{sec:cbm}

\begin{figure}[t]
    \centering
    \includegraphics[width=0.48\textwidth]{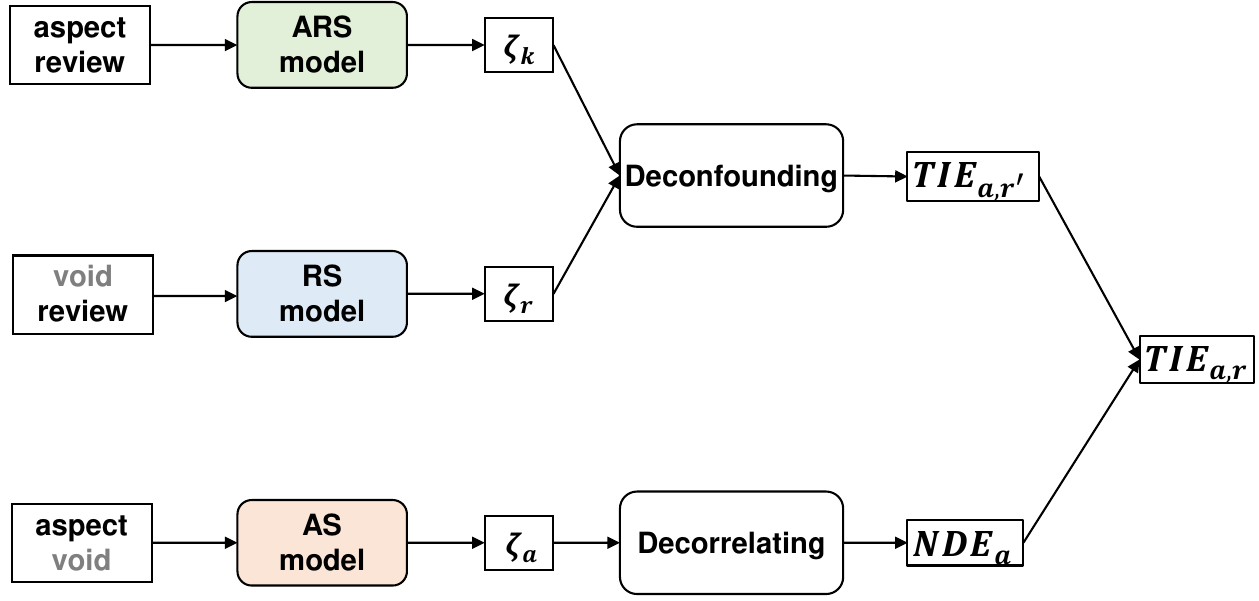}
    \caption{The framework of the proposed method.}
    \label{fig:framework}
\end{figure}

Based on Eq.~\eqref{eq:tie}, we can estimate each component and obtain an unbiased prediction.
However, as $R$ and $L$ are indirectly confounded with context $C$, it is not easy to calculate $L_{a,r,k}$ and $L_{a^*,r,k^*}$.
Therefore, we debias the review $R$ first.
\begin{equation}
    L_{a,r,k} = \Psi(\zeta_{a},\zeta_{r^{'}},\zeta_{k})
\end{equation}
where $\zeta_{k}$ denotes the logit of the softmax layer, $\Psi(\cdot)$ denotes the fusion function, specially $\zeta_{r^{'}}$ denotes the debiased output based on $R$.

There are mainly three types of causal intervention methods based on causal inference, the backdoor adjustment, the front-door adjustment, and the instrument variable adjustment. 
However, the front-door adjustment requires a mediator variable between input and output, which is not applicable in our SCM~\cite{zhang2024causal,aaai}. 
The instrument variable adjustment involves building up an extra instrument variable in SCM, which makes the already complex SCM even more complex~\cite{wang-etal-2022-causal}. 
So we choose the backdoor adjustment for debiasing the review branch.
Consider the SCM only contains $R$, $C$, and $L$, $C$ satisfies the backdoor criterion, and we can have:
\begin{equation}
\begin{split}
    P(L|do(R)) &= \sum_{\boldsymbol{c}}{P(L|R,C)P(C)} \\
    &= \sum_{\boldsymbol{c}}\frac{P(L, R|C)P(C)}{P(R|C)}
\label{ipw}
\end{split}
\end{equation}
where the $do(R)$ operator denotes a causal intervention that severs the direct effect of $R$ on $L$.

A common workaround is the application of the Normalized Weighted Geometric Mean (NWGM)~\cite{xu2015show} to approximate the effects of the $do$-operator.
Our approach adopts an Inverse Probability Weighting (IPW)~\cite{pearl2009causal} perspective, which provides a novel lens through which to approximate the infinite sampling of $(l, r)|c$ as shown in Eq.~\eqref{ipw}.

In a finite dataset, the observable instances of $(l, r)$ for each unique $c$ are limited. 
Consequently, the number of $c$ values considered in our equation equates to the available data samples rather than the theoretically infinite possibilities of $c$.
Backdoor adjustment bridges the gap between the confounded and de-confounded models, allowing us to treat samples from the confounded model as if they were drawn from the de-confounded scenario.
This leads to an approximation:
\begin{equation}
\begin{split}
P(L|do(R=r)) &\approx \widetilde{P}(L, R|C=c)\\
    &\approx \frac{1}{K}\sum_{k=1}^{K} \widetilde{P}(L, R=r^k|C=c)
\label{equ3}
\end{split}
\end{equation}
where $\widetilde{P}$ denotes the inverse weighted probability. 
We employ a multi-head strategy, inspired by \citet{vaswani2017attention} to refine the granularity of our sampling by partitioning the weight and feature dimensions into $K$ groups, $r_k$ denotes the review information in the group $K=k$.
For simplicity, subsequent discussions will omit $C = c$, though it is understood that $r$ remains dependent on $c$.
We will employ $\TDE$ to debias this effect following \citet{Liu_2022_CVPR} and \citet{tang2020long}.

The energy-based model~\cite{lecun2006tutorial} framework underpins our modeling of $\widetilde{P}$, where the softmax-activated probability is proportional to an energy function defined as:
\begin{equation}
\begin{split}
    \widetilde{P}(L=l, R=r^k) \propto E(l,r^k;w^k) \\
    =  \tau\frac{f(l,r^k;w^k)}{g(l,r^k;w^k)}
\end{split}
\end{equation}
with $\tau$ serving as a scaling factor analogous to the inverse temperature in Gibbs distributions~\cite{geman1984stochastic}, $w^k$ 
denotes the weight parameter in the group $K=k$.
The numerator $f(l,r^k;w^k)$ represents the unnormalized effect, calculated as logits $(w^k)^\top r^k$, while the denominator $g(l,r^k;w^k)$ serves as a normalization term(or propensity score~\cite{austin2011introduction}), which ensures balanced magnitudes of the variables.
The denominator, \ie, inverse probability weight, becomes the propensity score under the energy-based model, where the effect is divided into the controlled group $\lVert w^k \rVert \cdot \lVert r^k \rVert$ and the uncontrolled group $\epsilon \cdot \lVert r^k \rVert$.

The computation of logits for $P(L|do(R=r))$ is thus expressed as:
\begin{align}
    P(L|do(R)) 
    &= \frac{\tau}{K}  \sum_{k=1}^{K} \frac{(w^k)^\top r^k}{(\lVert w^k \rVert + \epsilon)\lVert r^k \rVert}
\label{equ:3}
\end{align}

Now we need to obtain context features given the current samples to force the model to concentrate on the debiased review based on $\TDE$.
We assume $U$ as a confounder set $\{u_i\}^N_{i=1}$, where $N$ is the number of aspects in dataset and
$u_i$ is the prototype for the context of class $i$ in feature space.
Review features can be linearly or non-linearly represented by the manifolds~\cite{turk1991face,candes2011robust}, and so are the context features.
Therefore, we model the review-specific context features $C$ of current samples as follows:
\begin{equation}
C = f(r, U) = \sum_{N=1}^{N}P(u_{n}|r)u_{n}
\end{equation}
where $P(u_{i}|r)$ is the classification probability of the feature $r$ belonging to the context of class $i$.

The last remaining difficulty is implementing the contextual confounder set $U$. 
To obtain more useful contextual information, we employ the lower  $\mathcal{K}$ layers of the model on $R \rightarrow L$ branch, which is in early training to model $U$.
It is motivated by three primary considerations: 
\textbf{First}, the acknowledgment of the intrinsic wealth of contextual semantic information harbored within pre-trained language models ~\cite{liu2019roberta,devlin-etal-2019-bert} due to their extensive pre-training.
\textbf{Second}, our requirement is not for highly advanced semantics but rather for contextual information~\cite{zeiler2014visualizing,liu2022contextual};
previous empirical studies~\cite{jawahar-etal-2019-bert,liu-etal-2019-linguistic,geva-etal-2021-transformer} have shown that encoder-only models exhibit superior performance in capturing contextual information at lower layers. 
\textbf{Third}, in the initial stages of training, the model's classification capabilities predominantly rely on context.

To be specific, we encode each $r$ using the aforementioned method, and if $r$ contains a specific aspect, it is then represented as the representation of the corresponding $u_n$, and we apply the mean feature as the final $u_n$ representation.

Given the modeling of $U$ and $C$, we are ready for the representations of context bias. 
We model them as $r_c = \mathcal{F}(r, C)$. 
Following~\citet{liu2022contextual}, we choose $W\cdot concat(r, M)$ to map since adding more networks to learn how much we need from the context is better.

Now we can debias the impact of $C$ on $R$ ~($C \rightarrow R$) based on $\TDE$. 
The final definition of debiased $r^{'}$ is as follows:
\begin{align}
    \zeta_{r^{'}}  = \frac{\tau}{K}  \sum_{k=1}^{K}  \frac{(w^k)^\top }{(\lVert w^k \rVert + \epsilon)}
    \left( \frac{r^k}{\lVert r^k \rVert}  -\frac{r^k_{c}}{\lVert r^k_{c}\rVert} \right)
\label{yr_calculate}
\end{align}

\subsection{Decorrelating the Aspect Branch with Counterfactual Reasoning}

While we have successfully mitigated contextual bias in the $R \rightarrow L$ pathway, the ABSA model, as delineated in Figure~\ref{fig:SCM}, remains susceptible to aspect-only bias. 
This bias persists because the prediction, denoted as $L_{a,r^{'},k}$, is directly influenced by the aspect variable $A$. 
To address this, we introduce a counterfactual reasoning approach that estimates the direct causal effect of $A$ on $L$, effectively isolating the influence of $R$ and $K$.
Figure~\ref{fig:framework} shows the causal graph of the counterfactual world for ABSA which describes the scenario when $A$ is set to different values $a$ and $a^*$.
We also set $R$ to its reference value $r^*$, therefore $K$ would attain the value $k^*$ when $R = r^*$ and $A = a^*$.
In this way,the inputs of $R$ and $K$ are blocked, and the model
can only rely on the given aspect $a$ for detection.
The natural direct effect ($\NDE$) of $A$ on $L$, which represents the aspect-only bias, is calculated as follows:
\begin{equation}
\NDE_{a} = L_{a,r^*,k^*} - L_{a^*,r^*,k^*}
\end{equation}
To eliminate this bias, we adjust  $\TE$ by subtracting  $\NDE$, yielding $\TIE$ in Eq.~\eqref{eq:tie}.

Following the previous studies, we calculate the prediction $L_{a,r,k}$ through a model ensemble with a fusion function:
\begin{align}
L_{a,r,k} &= L(A=a, R=r^{'}, K=k) \nonumber\\
&= \Psi(\zeta_{a},\zeta_{r^{'}},\zeta_{k}) 
\nonumber \\
    &= \zeta_{k} + \tanh(\zeta_{a}) + \tanh(\zeta_{r^{'}}) 
\label{eq:pred}
\end{align}
where $\zeta_{r^{'}}$ is the output of the review-only branch ($\ie$,
$R \rightarrow L$ ), $\zeta_{a}$ is the output of the aspect-only branch
($\ie$, $ A \rightarrow L$ ), and $\zeta_{k}$ is the output of fused
features branch ($\ie$, $K \rightarrow L$ ) as shown in Figure~\ref{fig:framework}.
$\TIE$ is the debiased result we used for inference.

\subsection{Training and Inference}
We compute separate losses for each branch during the training stage in line with the methodologies adopted by recent studies ~\cite{wang2021clicks,niu2021counterfactual,tian2022debiasing,chen2023causal}.
These branches comprise the fused feature branch (base ABSA, $\mathcal{L}_{K}$), the aspect-only branch ($\mathcal{L}_{A}$), and the debiased review-only branch ($\mathcal{L}_{R}$).
The collective minimization of these losses forms a comprehensive multi-task training objective,  which serves to optimize the model parameters. 
The training objective is formally expressed as:
\begin{equation}
\mathcal{L}=\mathcal{L}_{K}+\alpha \mathcal{L}_{A}+\beta \mathcal{L}_{R}
\end{equation}
where $\alpha$ and  $\beta$ are hyperparameters that control the contribution of each branch to the overall training objective.

The loss component $\mathcal{L}_{K}$ corresponds to the cross-entropy loss calculated from the predictions of $\Psi(\zeta_{a}, \zeta_{r^{'}}, \zeta_{k})$, as defined in Eq.~\eqref{eq:pred}. 
Similarly, the aspect-only and debiased review-only losses are denoted as $\mathcal{L}_{A}$ and $\mathcal{L}_{R}$ respectively.

We use debiased $\TIE_{a,r}$ in Eq.~\ref{eq:tie} for inference.

\section{Experiments}

\begin{table*}[t]
\centering
\small
\begin{tabular}{lcccccc}
\toprule
      & \multicolumn{3}{c}{\textbf{Laptop}} & \multicolumn{3}{c}{\textbf{Restaurant}} \\
\midrule
      \textbf{Model}  & \textit{Acc.}  & F1-score & \texttt{ARS}    & \textit{Acc.}   & F1-score  & \texttt{ARS}   \\
\midrule
MemNet~\cite{tang-etal-2016-aspect}  & - & - & 16.93 & -  & - & 21.52 \\
GatedCNN~\cite{xue-li-2018-aspect} & -  & -  & 10.34  & - & - & 13.12 \\
AttLSTM~\cite{wang-etal-2016-attention} & - & - & 9.87 & - & - & 14.64 \\
TD-LSTM~\cite{tang-etal-2016-effective}  & -  & -  & 22.57 & - & -& 30.18 \\
GCN~\cite{zhang-etal-2019-aspect} & -  & - & 19.91 & - & - & 24.73 \\
\midrule
BERT-Sent~\cite{xing-etal-2020-tasty}   & -   & - & 14.70 & -  & - & 10.89  \\
CapsBERT~\cite{jiang-etal-2019-challenge} & -   & - & 25.86 & -  & - & 55.36  \\
BERT-PT~\cite{xu-etal-2019-bert}  & - & - & 53.29  & - & - & 59.29  \\
GraphMerge~\cite{hou2021graph} & - & - & 52.90  & - & - & 57.46  \\
NADS~\cite{cao2022aspect}  & - & - & 58.77  & - & - & 64.55  \\
SENTA~\cite{bi2021interventional}  & 67.23 & - & -  & 77.30 & - & -  \\
PT-SENTA~\cite{bi2021interventional}  & 74.16 & - & -  & 80.91 & - & -  \\
ChatGPT~\cite{wang2023chatgpt} & 68.89 & 56.22 & 46.39  & 79.21 & 61.33 & 45.01  \\ \midrule

BERT~\cite{xing-etal-2020-tasty}   & - & - & 50.94  & - & - & 54.82  \\
BERT$^\dag$ & 70.43 & 66.55 & 49.53  & 78.56 & 69.35 & 57.86  \\
\rowcolor{mygray} \textbf{DINER}(BERT-based)& 72.56 & 68.40 & 53.76 & 80.69& 72.79 & 62.23 \\
\midrule
RoBERTa~\cite{ma-etal-2021-exploiting} & 73.57 & 69.26 & -  & 79.08 & 72.79 & -  \\
RoBERTa$^\dag$ & 74.96 & 72.16 & 56.27  & 79.26 & 70.47 & 59.96 \\
\rowcolor{mygray} \textbf{DINER}(RoBERTa-based)& \textbf{76.51} & \textbf{73.27} & \textbf{59.40}  & \textbf{82.46} & \textbf{76.92} & \textbf{64.02}  \\
\bottomrule
\end{tabular}
\caption{We retrained BERT$^\dag$, RoBERTa$^\dag$  as fair baselines ensuring that comparisons are made under similar training settings, which is crucial for validating \textbf{DINER}'s superior performance.}
\label{tab:main_result}
\end{table*}

\subsection{Datasets}
We conduct training on the original SemEval 2014 Laptop and Restaurant datasets~\cite{pontiki-etal-2014-semeval}, and perform testing on the ARTS datasets, as introduced by ~\citet{xing-etal-2020-tasty}, to assess the efficacy of the proposed method.
Detailed information about the ARTS datasets is shown in Appendix~\ref{sec:example}.

\subsection{Baselines}
We consider baselines in the ARTS original paper~\cite{xing-etal-2020-tasty}, which are listed in  Appendix~\ref{sec:baseline} and following strong baselines for comparison:\\
\textbf{GraphMerge:}~\citet{hou2021graph} combine multiple dependency trees using a graph-ensemble technique for aspect-level sentiment analysis.\\
\textbf{SENTA:}~\citet{bi2021interventional} propose a novel Sentiment Adjustment model, employing backdoor adjustment to mitigate confounding effects.
And \textbf{PT-SENTA} use BERT-PT~\cite{xu-etal-2019-bert} as backbone.\\
\textbf{NADS:}~\citet{cao2022aspect} apply no-aspect contrastive learning to reduce aspect sentiment bias and improve sentence representations.\\
\textbf{ChatGPT:} ChatGPT is a conversational version of GPT-3.5 model~\cite{ouyang2022training,chatgpt}. 
We use the \texttt{gpt-3.5-turbo-0125} API from OpenAI\footnote{\url{https://platform.openai.com/docs/models/gpt-3-5}}.
The prompts for this task are presented in Appendix~\ref{sec:prompt}.

\subsection{Implementations}
Our method is model-agnostic. In the empirical study, we utilize two types of mainstream encoder-only model, RoBERTa~\cite{liu2019roberta}\footnote{\url{https://huggingface.co/FacebookAI/roberta-base}} and BERT~\cite{devlin-etal-2019-bert}\footnote{\url{https://huggingface.co/bert-base-uncased}} as the backbone for our experiments.
For comprehensive details on the hyperparameters employed in our experiments, refer to Appendix~\ref{appendix:para}.

\subsection{Evaluation}
Following the previous works~\cite{wang-etal-2016-attention,xue-li-2018-aspect,cao2022aspect}, Accuracy (\textit{Acc.}), F1-score and Aspect Robustness Score (\texttt{ARS})~\cite{xing-etal-2020-tasty} are employed as complementary evaluation metrics. 
\texttt{ARS} considers the accurate classification of a source example and all its derived variants, produced through the aforementioned three strategies, as a single instance of correctness.

\section{Result and Analysis}
Table~\ref{tab:main_result} presents a detailed comparison of various models' performance for laptop and restaurant domains of the ARTS datasets, focusing on three key evaluation metrics: \textit{Acc.}, F1-score, and \texttt{ARS}.

Overall, PLMs, on average, perform better than non-PLMs due to the pre-trained knowledge and tasks, making them more robust.
Surprisely, ChatGPT does not get perform well in this task, exhibiting \texttt{ARS} scores of only 50.94 in the laptop domain and 54.82 in the restaurant domain, which are even lower than those of most PLMs in Table~\ref{tab:main_result}.
This underscores ChatGPT's relatively poor robustness in ARTS's variations, despite its otherwise robust performance across various other NLP tasks.

We evaluate \textbf{DINER} in two backbones: BERT-based and RoBERTa-based. 
These configurations are set to evaluate the effectiveness of \textbf{DINER} when integrated with different encoder-only PLMs.
And \textbf{DINER} based on RoBERTa tends to outperform its BERT counterparts, which may be attributed to RoBERTa's more robust pre-training on a larger and more diverse corpus, leading to better generalization capabilities~\cite{liu2019roberta}.
The results are compelling, showing that \textbf{DINER}(RoBERTa-based) model achieves the state-of-the-art performance across all metrics in both the laptop and restaurant domains, with a notable \textit{Acc.} of 76.51 and 82.46, F1-scores of 73.27 and 76.92, and \texttt{ARS} of 59.40 and 64.02, respectively.
\textbf{DINER}(RoBERTa-based) demonstrates superior performance in the Laptop domain, outpacing the baseline RoBERTa$^\dag$ by margins of \textbf{1.55, 1.11, and 3.13} in terms of \textit{Acc.}, F1-score, and \texttt{ARS} metrics, respectively.
In the Restaurant domain, the model further extends its lead, achieving improvements of \textbf{3.20, 6.45, and 4.06} in the same metrics.
Similarly, the \textbf{DINER}(BERT-based) exhibits empirical enhancements.

\subsection{More Detailed Result}
\begin{table*}[ht]
\centering
\small
\begin{tabular}{lccccc}
\toprule
 & & \textsc{RevTgt}  & \textsc{RevNon} & \textsc{AddDiff} & \textsc{Original}\\
\midrule
\multicolumn{1}{c}{\multirow{2}{*}{Laptop}}   &\texttt{Vanilla}  &   62.45   &    85.93    &    76.33  & 80.41 \\
&\textbf{DINER}& 65.02(\textcolor{red}{$\uparrow$ 4.12\%})& 86.67(\textcolor{red}{$\uparrow$ 0.86\%}) & 78.06(\textcolor{red}{$\uparrow$ 2.27\%})&81.19(\textcolor{red}{$\uparrow$ 0.97\%})\\
\midrule
\multicolumn{1}{c}{\multirow{2}{*}{Restaurant}} &\texttt{Vanilla}&  64.06  &     82.66   &  83.48   &85.18  \\
&\textbf{DINER}& 70.69(\textcolor{red}{$\uparrow$ 10.35\%}) & 83.56(\textcolor{red}{$\uparrow$ 1.08\%}) & 86.07(\textcolor{red}{$\uparrow$ 3.10\%}) &87.32(\textcolor{red}{$\uparrow$ 2.51\%})\\
\bottomrule 
\end{tabular}
\caption{We use RoBERTa as the backbone.
\texttt{Vanilla} refers to RoBERTa$^\dag$ in Table~\ref{tab:main_result}.
We compare the \textit{Acc.} on the \texttt{Vanilla} and our \textbf{DINER} framework.
We also calculate the \textcolor{red}{\textit{change}} of accuracy.}
\label{table:detailed_result}
\end{table*}

We list in detail the performance of each model on the aforementioned three subsets of the ARTS datasets in Table~\ref{table:detailed_result}.
% Restaurant domain get a significant boost, and the fact that the test set data in the restaurant domain is inherently more challenging~\cite{xing-etal-2020-tasty} demonstrates the effectiveness of our approach.
In the Laptop domain, the baseline model RoBERTa$^\dag$ exhibits a \textsc{RevTgt} accuracy of 62.45, as  \textsc{RevTgt} is the most challenging subset. It requires the model to pay precise attention to the target sentiment words.
In contrast, the DINER framework significantly enhances this metric to 65.02, marking a \textcolor{red}{4.12\%} increment.
Similarly, for \textsc{RevNon} and \textsc{AddDiff}, DINER outperforms the Vanilla baseline with modest improvements of \textcolor{red}{0.86\%} and \textcolor{red}{2.27\%}, respectively.
The Restaurant domain further underscores the efficacy of the DINER framework, where a remarkable \textcolor{red}{10.35\%} improvement is observed in the \textsc{RevTgt} task, elevating the accuracy from 64.06 to 70.69. The framework also exhibits gains in \textsc{RevNon} and \textsc{AddDiff} tasks by \textcolor{red}{1.08\%} and \textcolor{red}{3.10\%,} respectively.
The significant improvement observed in the restaurant domain underscores the effectiveness of our methods, particularly given the inherently challenging nature of the test set data in this domain, as highlighted by~\citet{xing-etal-2020-tasty}.
Specifically, the more challenging the dataset, the greater the improvement our framework offers.
Interestingly, our method also gives a slight improvement on the \textsc{Original} test set, illustrating the fact that we have also de-biased the robust test data on the \textsc{Original} test set.

\subsection{Effects of the Two Branches in \textbf{DINER}}
\begin{table}[]
\centering
\small
\begin{tabular}{lcc}
\toprule
Methods & \textbf{Laptop}  & \textbf{Restaurant}\\
\midrule
\texttt{Vanilla}    &   74.96  &   79.26   \\
\midrule
$R \rightarrow L$ branch \\
\quad + Causal Intervention(NWGM)  &   75.44     &    81.02  \\
\quad + Causal Intervention(IPW)  & 75.50  &    81.19  \\
\quad \quad +$\TDE$  &    75.92    &    81.78 \\
\midrule
$A \rightarrow L$ branch \\
\quad + Counterfactual Inference  &    75.23    &   80.51   \\
\midrule
% CCD &     75.97   &     81.98 \\
% \midrule
\textbf{DINER} &     \textbf{76.51}   &  \textbf{82.46}    \\
\bottomrule 
\end{tabular}
\caption{Ablation studies on two branches of our method. Experiments are based on RoBERTa backbone, \textit{Acc.} are reported.}
\label{table:ablation}
\end{table}

We delve into the empirical evaluation of the dual-branch architecture underpinning the DINER framework, specifically examining its constituent elements through ablation studies. 
The studies are shown in Table~\ref{table:ablation}, offering insights into the incremental benefits conferred by each branch.

For the $R \rightarrow L$ branch, NWGM~\cite{xu2015show} yields a marginal improvement in accuracy across both domains. 
The method of IPW~\cite{pearl2009causal} further enhances performance, suggesting the efficacy of backdoor adjustment intervention in the methods, and IPW has a more precise approximation compared to NWGM~\cite{xu2015show}.
We further debias context based on $\TDE$, as described in Section~\ref{sec:cbm}, and performance is further enhanced upon the application of Counterfactual Reasoning.

Parallel to this, the $A \rightarrow L$ branch investigates the impact of Counterfactual Inference.
After conducting Counterfactual Inference at this branch, \textit{Acc.} in the Laptop and Restaurant domains improved by 0.27 and 1.25, respectively.
In the Restaurant domain, the bias associated with aspects is more pronounced.

By comparing the performance improvements at both branches, we can also discern that the bias and shortcuts from $R \rightarrow L$ branch are more pronounced, and our approach has effectively addressed these issues.

% CCD framework~\cite{chen2023causal} are used to fake news detection, it conduct
% causal interventions via backdoor adjustment to remove spurious correlations from sentence branch and apply counterfactual reasoning to address image-only bias.
% And it apply NWGM~\cite{xu2015show}  to approximate the effects of the $do$-operator.
% It can transfer to our framework seamlessly, which is a very strong baseline.

\subsection{Impact of Different Fusion Strategies}

\begin{table}[]
\centering
\small
\begin{tabular}{lcc}
\toprule
Fusion Strategy  & \textbf{Laptop}  & \textbf{Restaurant}\\
\midrule
 \texttt{MUL}-\texttt{Vanilla} & 53.01  & 65.52 \\
 \texttt{MUL}-\texttt{sigmoid} & 63.72 & 76.35\\
 \texttt{MUL}-$\tanh$ & 52.10 &61.36 \\
 \texttt{SUM}-\texttt{Vanilla} &74.32 & 80.14\\
 \texttt{SUM}-\texttt{sigmoid} & 75.97 & 81.76\\
 \texttt{SUM}-$\tanh$ & \textbf{76.51}&\textbf{82.46} \\
\bottomrule 
\end{tabular}
\caption{Impact of Different Fusion Strategies.}
\label{table:fusion}
\end{table}

\begin{table*}[t]
\centering
\small
\begin{tabularx}{\textwidth}{lXlll}
\toprule
\textbf{Type} & \textbf{Examples(Target Aspect: food)}  & \textbf{Gold}  & \textbf{Baseline} & \textbf{DINER} \\
\midrule
\multirow{2}{*}{\textsc{Original}} & The \textbf{food} is top notch, the service is attentive, and the atmosphere is great.&  \multirow{2}{*}{Positive} & \multirow{2}{*}{Positive \yes} & \multirow{2}{*}{Positive \yes} \\
\multirow{2}{*}{\textsc{RevTgt}} & The \textbf{food} is \underline{nasty, but} the service is attentive, and the atmosphere is great. & \multirow{2}{*}{Negative} & \multirow{2}{*}{Negative \yes} & \multirow{2}{*}{Negative \yes} \\
\multirow{2}{*}{\textsc{RevNon}} & The \textbf{food} is top notch, the service is \underline{heedless}, \underline{but} the atmosphere is \underline{not great}. & \multirow{2}{*}{Positive} & \multirow{2}{*}{Negative \no} & \multirow{2}{*}{Positive \yes} \\
\multirow{2}{*}{\textsc{AddDiff}} & The \textbf{food} is top notch, the service is attentive, and the atmosphere is great, \underline{but music is too heavy, waiters is angry and staff is arrogant.}& \multirow{2}{*}{Positive} & \multirow{2}{*}{Negative \no} & \multirow{2}{*}{Positive \yes} \\
\bottomrule 
\end{tabularx}
\caption{Examples of case study. The corresponding gold labels and the predictions for each example are presented. }
\label{table:casestudy}
\end{table*}

Following prior studies~\cite{wang2021clicks,niu2021counterfactual,chen2023causal}, 
we devise several differentiable arithmetic binary operations for the fusion strategy in Eq.~\eqref{eq:fusion}:
\begin{equation}
    \left\{ 
    \begin{array}{l}
    \text{\texttt{MUL}-\texttt{Vanilla}} : L_{a,r^{'},k} = \zeta_{a} \cdot \zeta_{r^{'}} \cdot \zeta_k, \\
    \text{\texttt{MUL}-\texttt{sigmoid}}: L_{a,r^{'},k} = \zeta_{k} \cdot \sigma(\zeta_a) \cdot \sigma(\zeta_{r^{'}}), \\
    \text{\texttt{MUL}-$\tanh$}: L_{a,r^{'},k} = \zeta_k \cdot \tanh(\zeta_a) \cdot \tanh(\zeta_{r^{'}}), \\
    \text{\texttt{SUM}-\texttt{Vanilla}}: L_{a,r^{'},k} = \zeta_{a} + \zeta_{r^{'}} + \zeta_k, \\
    \text{\texttt{SUM}-\texttt{sigmoid}}: L_{a,r^{'},k} = \zeta_k + \sigma(\zeta_a) + \sigma(\zeta_{r^{'}}), \\
    \text{\texttt{SUM}-$\tanh$}: L_{a,r^{'},k} = \zeta_k + \tanh(\zeta_a) + \tanh(\zeta_{r^{'}}) 
    \end{array} 
    \right.
    \label{eq:fusion}
\end{equation}
The \textit{Acc.} performance of six distinct different fusion strategies are
reported in Table~\ref{table:fusion}.
From the table, we can find
that the \texttt{MUL} fusion, regardless of the activation function, consistently underperforms in comparison to its \texttt{SUM} counterparts.
Apparently, \text{\texttt{SUM}} fusion strategies are more stable and robust, and more suitable for the ASBA task.
The superior performance of \texttt{SUM} fusion strategies, particularly with the $\tanh$ activation, underscores the effectiveness of the additive strategy in capturing the nuanced interplay of features pertinent to the ABSA task.
% \text{\texttt{SUM}-$\tanh$} achieves the best performance over
% the other fusion strategies.

\subsection{Case Study}

To demonstrate the efficacy of the proposed method, we present a case study featuring a sample and its three adversarial variants in Table~\ref{table:casestudy}.
We compare our proposed method based on RoBERTa with the baseline RoBERTa$^\dag$.

From the table, the results clearly demonstrate that our method \textbf{DINER}, exhibits enhanced robustness compared to the baseline approach. 
Specifically, \textsc{Original} and \textsc{RevTgt} types, where either no changes or direct negative changes were made to the targeted aspect, both methods perform equally well. 

However, the distinction in performance is evident in more complex adversarial examples.
In the \textsc{RevNon} type, where distractors are introduced in non-target aspects (\eg, service and atmosphere), the baseline fails to maintain its accuracy, misclassifying the overall sentiment as Negative. 
In contrast, DINER successfully recognizes the sentiment as Positive, reflecting its ability to isolate the influence of perturbations to non-target aspects.
The \textsc{AddDiff} type further complicates the scenario by adding multiple negative aspects unrelated to the target.
Despite these challenges, DINER continues to accurately assess the sentiment towards the food as Positive, whereas the baseline erroneously shifts to a Negative prediction.

The resilience of our method to adversarial conditions suggests it is well-suited for real-world environments where reliable sentiment analysis is crucial.

\section{Conclusion}
In this paper, to debias the target and review in ABSA simultaneously, a novel debiasing framework, DINER, is proposed with multi-variable causal inference. 
Specifically, the aspect is assumed to have a direct correlation with the label, so a counterfactual reasoning-based intervention is employed to debias the aspect branch.
In the meantime, the sentiment words towards the target in the review are assumed to be indirectly confounded with the context, where a backdoor adjustment-based intervention is employed to debias the review branch.
Extensive experiments show the effectiveness of the proposed method in debiasing ABSA compared to normal state-of-the-art ABSA methods and debiasing methods.

\section*{Limitations}
Though achieving promising results in the experiments, our work still has the following limitations.
\begin{itemize}[leftmargin=*]
    \item Though the proposed method is based on multi-variable causal inference, the causal effects of the target aspect and the review are assumed to be independent, which means no interaction between the target and the review is modeled or considered.
    \item The proposed method is only evaluated on two robustness testing datasets for ABSA. More real-world datasets and more data transformation methods should be evaluated for future work.
    \item The general ABSA task includes the joint extraction of aspect and sentiment polarity, while the proposed method restricts the task to a given aspect. 
    Future work should be considered for more generalized ABSA tasks.
\end{itemize}

\section*{Acknowledgement}
The authors would like to thank the anonymous reviewers for their insightful comments. This work is funded by the National Natural Science Foundation of China (62176053). This work is supported by the Big Data Computing Center of Southeast University.
% Entries for the entire Anthology, followed by custom entries
\bibliography{anthology,custom}
\bibliographystyle{acl_natbib}

\clearpage
\newpage
\appendix
\section{Dataset Example}
\begin{table}[!t]
\centering
\small
\begin{tabular}{lccccc}
\toprule
 & \textsc{RevTgt}  & \textsc{RevNon} & \textsc{AddDiff}  & ALL \\
\midrule
Laptop     &     466   &  135      &    638    &   1877  \\
Restaurant &     846   &  444      &    1120   &   3530  \\
\bottomrule 
\end{tabular}
\caption{The statistics of datasets being evaluated.}
\label{table:sta}
\end{table}

ARTS datasets employ three distinct strategies to rigorously test the model's robustness: 
\textsc{RevTgt} is to generate sentences that reverse the original sentiment of the target aspect.
\textsc{RevNon} is to change the target sentiment.
\textsc{AddDiff} investigate if adding more nontarget aspects can confuse the model.
We provide concrete instances of how each strategy is applied to manipulate aspect sentiment within the dataset in Table~\ref{table:example-ARTS}.
Detailed statistics of the test sets are provided in Table~\ref{table:sta}.
\label{sec:example}
\begin{table*}[!t]
\small
\centering
\begin{tabular}{lc}
\toprule
Type & Review \\
\midrule
\textsc{Original}     &   Tasty \textbf{burgers}, and crispy fries. (Target Aspect: \textbf{burgers})   \\
\textsc{RevTgt}  &   \underline{Terrible} \textbf{burgers}, but crispy fries.   \\
\textsc{RevNon} &   Tasty \textbf{burgers}, but \underline{soggy} fries.   \\
\textsc{AddDiff} &  Tasty \textbf{burgers}, crispy fries, \underline{but poorest service ever!}  \\
\bottomrule 
\end{tabular}
\caption{The adversarial examples of the original sentence. 
Each example is annotated with the \textbf{Target Aspect}, and \underline{altered sentence parts}.}
\label{table:example-ARTS}
\end{table*}

\section{Baselines}
\label{sec:baseline}
\textbf{TD-LSTM:}~\citet{tang-etal-2016-effective} use dual LSTMs to encode context around a target aspect, combining final states for sentiment classification.\\
\textbf{AttLSTM:}~\citet{wang-etal-2016-attention} introduce an Attention-based LSTM that merges aspect and word embeddings for each token.\\
\textbf{GatedCNN:}~\citet{xue-li-2018-aspect} utilize a Gated CNN with a Tanh-ReLU mechanism, integrating aspect embeddings with CNN-encoded text.\\
\textbf{MemNet:} ~\citet{tang-etal-2016-aspect} employ memory networks, using sentences as external memory to compute attention based on the target aspect.\\
\textbf{GCN:}~\citet{zhang-etal-2019-aspect} apply a GCN to the sentence's syntax tree, followed by an aspect-specific masking layer.\\
\textbf{BERT:}~\citet{xu-etal-2019-bert} use a BERT-based~\cite{devlin-etal-2019-bert} baseline and takes as input the concatenation of the aspect and the review.\\
\textbf{BERT-Sent:}~\cite{xu-etal-2019-bert} BERT-Sent takes as input reviews without aspect.\\
\textbf{BERT-PT:} ~\citet{xu-etal-2019-bert} enhance BERT's capabilities through post-training on additional review datasets.\\
\textbf{CapsBERT:}~\citet{jiang-etal-2019-challenge}  use BERT to encode sentences and aspect terms, then utilize Capsule Networks for polarity prediction.

\section{ChatGPT Prompt}
\label{sec:prompt}
We conduct 3-shot prompting experiments on the ARTS datasets following~\cite{wang2023chatgpt}.
We set the decoding temperature as 0 to increase ChatGPT’s determinism.
The prompts are presented in Table~\ref{table:prompt}.
\begin{table*}[]\centering
\small
\begin{tabular}{lp{14cm}}\toprule
\textbf{Dataset} & \textbf{Prompt} \\
\midrule
Laptop & \makecell[lp{14cm}]{
Sentence: The \textbf{screen} almost looked like a barcode when it froze.
\\
What is the sentiment polarity of the aspect \textbf{screen} in this sentence?\\
Label: negative\\
Sentence: Screen, keyboard, and mouse: If you cant see yourself spending the extra money to jump up to a Mac the beautiful screen, responsive \textbf{island backlit keyboard}, and fun multi-touch mouse is worth the extra money to me alone. \\
What is the sentiment polarity of the aspect \textbf{island backlit keyboard} in this sentence?\\
Label: positive\\
Sentence: Size: I know 13 is small (especially for a desktop replacement) but with an \textbf{external monitor}, who cares.
\\
What is the sentiment polarity of the aspect \textbf{external monitor} in this sentence?\\
Label: neutral\\
Sentence: \textcolor{red}{$\{$sentence$\}$} \\
What is the sentiment polarity of the \textcolor{blue}{$\{$aspect$\}$} in this sentence?}
 \\
 \midrule
 Restaurant &\makecell[lp{14cm}]{
 Sentence: Our \textbf{server} was very helpful and friendly.
\\
What is the sentiment polarity of the aspect \textbf{server} in this sentence?\\
Label: positive\\
Sentence: We had \textbf{reservations} at 9pm, but was not seated until 10:15pm. \\
What is the sentiment polarity of the aspect \textbf{reservation} in this sentence?\\
Label: negative\\
Sentence: It's the perfect restaurant for NY life style, it got cool design, awsome drinks and food and lot's of good looking people eating and hanging at the pink \textbf{bar}...
\\
What is the sentiment polarity of the aspect \textbf{bar} in this sentence?\\
Label: neutral\\
Sentence: \textcolor{red}{$\{$sentence$\}$} \\
What is the sentiment polarity of the \textcolor{blue}{$\{$aspect$\}$} in this sentence?}
 \\
\bottomrule
\end{tabular}
\caption{The prompts used for prompting ChatGPT for each domain.}
\label{table:prompt}
\end{table*}

\section{Model Hyper Parameters}\label{appendix:para}
The model parameters are optimized by AdamW~\cite{loshchilov2018decoupled}, with a learning rate of 5e-5 and weight decay of 0.01. The batch size is 256, and a dropout probability of 0.1 is used. 
The number of training epochs is 20. 
We explore the hyperparameters $\alpha$ and  $\beta$, setting their values to $\{$0.6, 0.8, 1, 1.2, 1.4$\}$ for each, respectively.
The optimal values for $\alpha$ and $\beta$ are 0.8 and 1.0, respectively.
We set $\mathcal{K}$ in set $\{$3,6,9$\}$ in accordance with the theoretical principles discussed in~\cite{geva-etal-2021-transformer}.
Our implementation leverages the \textit{PyTorch}\footnote{\url{https://github.com/pytorch/pytorch}} framework and \textit{HuggingFace Transformers}\footnote{\url{https://github.com/huggingface/transformers}} library~\cite{wolf-etal-2020-transformers}.
Our experiments are carried out with an NVIDIA A100 80GB GPU.

\end{document}